\pdfoutput=1

\documentclass[11pt]{article}

\usepackage[]{EACL2023}

\usepackage{times}
\usepackage{latexsym}
\usepackage{amsmath}
\usepackage{amssymb}
\usepackage{tikz}
\usepackage{ctable}
\usepackage{graphicx}
\usepackage{caption}
\usepackage{balance}
\usepackage{subfig}
\usepackage{multirow}

\usepackage[justification=centering]{caption}
\usepackage[T1]{fontenc}

\usepackage[utf8]{inputenc}
\usepackage{tabularx}
\usepackage{tabulary}
\usepackage{textcomp}
\usepackage{booktabs}
\usepackage{colortbl}
\usepackage{microtype}
\title{Slot Induction via Pre-trained Language Model Probing and \\ Multi-level Contrastive Learning}

\author{Hoang H. Nguyen$^{1}$, Chenwei Zhang$^{2}$, Ye Liu$^{3}$, Philip S. Yu$^{1}$\\
  $^1$ Department of Computer Science, University of Illinois at Chicago, Chicago, IL, USA \\
  $^2$ Amazon, Seattle, WA, USA \\
  $^3$ Salesforce Research, Palo Alto, CA, USA \\
  \texttt{\{hnguy7,psyu\}@uic.edu, cwzhang@amazon.com, yeliu@salesforce.com}
  }

\begin{document}
\maketitle
\begin{abstract}
Recent advanced methods in Natural Language Understanding for Task-oriented Dialogue (TOD) Systems ~(e.g., intent detection and slot filling) require a large amount of annotated data to achieve competitive performance. In reality, token-level annotations (slot labels) are time-consuming and difficult to acquire. In this work, we study the Slot Induction~(SI) task whose objective is to induce slot boundaries without explicit knowledge of token-level slot annotations. We propose leveraging Unsupervised Pre-trained Language Model~(PLM) Probing and Contrastive Learning mechanism to exploit (1) unsupervised semantic knowledge extracted from PLM, and (2) additional sentence-level intent label signals available from TOD. Our approach is shown to be effective in SI task and capable of bridging the gaps with token-level supervised models on two NLU benchmark datasets. When generalized to emerging intents, our SI objectives also provide enhanced slot label representations, leading to improved performance on the Slot Filling tasks. \footnote{Our code and datasets are publicly available at \href{https://github.com/nhhoang96/MultiCL\_Slot\_Induction}{https://github.com/nhhoang96/MultiCL\_Slot\_Induction}}
\end{abstract}

\section{Introduction}
Natural Language Understanding~(NLU) has become a crucial component of the Task-oriented Dialogue (TOD) Systems. The goal of NLU is to extract and capture semantics from users' utterances \footnote{In our work, we use the term \textbf{utterance} and \textbf{sentence} interchangeably.}. There are two major tasks in NLU framework, including intent detection~(ID) and slot filling~(SF)~\cite{tur2011spoken}. While the former focuses on identifying overall users' intents, the latter extracts semantic concepts from natural language sentences. In NLU tasks, intents denote sentence-level annotations while slot types represent token-level labels. 



Despite recent advances, state-of-the-art NLU methods~\cite{haihong2019novel,goo2018slot} require a large amount of annotated data to achieve competitive performance. However, the fact that annotations, especially token-level labels, are expensive and time-consuming to acquire severely inhibits the generalization capability of traditional NLU models in an open-world setting 
\citep{louvan2020recent,xia2020cg}. Recent works attempt at tackling the problems in low-resource settings on both intent level~\citep{xia2018zero,nguyen2020dynamic, siddique2021generalized} and slot level ~\citep{yu2021cross,glass2021robust}. However, 
most approaches remain restricted to closed-world settings where there exist pre-defined sets of seen and emerging sets of classes. Some approaches even require additional knowledge from related token-level tasks that might not be readily available.

Additionally, with increasing exposure to the ever-growing number of intents and slots, TOD systems are expected to acquire task-oriented adaptation capability by leveraging both inherent semantic language understanding and task-specific knowledge to identify the crucial emerging concepts in the users' utterances. 
This ability can be referred to as \textbf{Slot Induction} in TOD Systems.

Recently, Pre-trained Contextualized Language Models ~(PLM) such as BERT~\cite{devlin2019bert} have shown promising capability of capturing semantic and syntactic structure without explicit linguistic pre-training objectives~\cite{jawahar2019does, rogers2020primer, wu2020perturbed}. Despite imperfections, the captured semantics from PLM via unsupervised probing mechanisms could be leveraged to induce important semantic phrases covering token-level slot labels.

Additionally, as an effective unsupervised representation learning mechanism \citep{wei2019eda,gao2021simcse}, Contrastive Learning~(CL) is capable of refining the imperfect PLM semantic phrases in a self-supervised manner to mitigate biases existent in the PLM. In specific, given a sample phrase \textit{in the same area} corresponding to \textit{spatial\_relation} slot type, as a presumed structural knowledge, PLM tends to split the preposition and determiner from the noun phrase during segmentation, resulting in \textit{in the} and \textit{same area}. Despite its structural correctness, the identified segments fail to align with ground truth slots due to the lack of knowledge from the overall utterance semantics.   

On the other hand, CL can also be leveraged on a sentence level when intent labels are available. In fact, there exist strong connections between slot and intent labels \citep{zhang2019joint, wu-etal-2020-slotrefine}.  For instance, utterances with \textit{book\_restaurant} intent tend to contain \textit{location} slots than those from \textit{rate\_book} intent. Therefore, as intent labels are less expensive to acquire, they could provide additional signals for CL to induce slot labels more effectively when available.

In this work, we propose leveraging PLM probing together with CL objectives for Slot Induction~(SI) task. Despite imperfections, PLM-derived segmentations could produce substantial guidance for SI when slot labels are not readily available. We introduce CL to further refine PLM segmentations via (1) segment-level supervision from unsupervised PLM itself, and (2) sentence-level supervision from intent labels to exploit the semantic connections between slots and intents. Our refined BERT from SI objectives can produce effective slot representations, leading to improved performance in slot-related tasks when generalized towards emerging intents.

Our contributions can be summarized as follows:

\noindent$\bullet$ We propose leveraging semantic segments derived from Unsupervised PLM Probing~(UPL) to induce phrases covering token-level slot labels. We name the task as Slot Induction.

\noindent$\bullet$ We propose enhancing the quality of PLM segments with Contrastive Learning refinement to better exploit (1) unsupervised segment-level signals from PLM, (2) sentence-level signals from intent labels to improve SI performance.

\noindent$\bullet$ We showcase the effectiveness of our proposed SI framework and its ability to produce refined PLM representations for token-level slots when generalized to emerging intents.



\section{Related Work}
\vspace*{-0.1cm}
\paragraph{Pre-trained Language Model Probing} Pre-trained Language Models~(PLMs) have been shown to possess inherent syntactic and semantic information. Different probing techniques are developed to investigate the knowledge acquired by PLMs, either from output representations~\cite{wu2020perturbed}, intermediate representations~\cite{sun2019patient}, or attention mapping~\cite{clark2019does, yu2022unsupervised}. Unlike previous probing techniques that focus on deriving syntactic tree structure, we leverage semantically coherent segments recognized by PLMs to induce phrases containing token-level slot labels in NLU tasks for TOD Systems. 

\paragraph{Contrastive Learning}
Contrastive Learning (CL) has been widely leveraged as an effective representation learning mechanism~\cite{oord2018representation}. The goal of CL is to learn the discriminative features of instances via different augmentation methods. In Natural Language Processing~(NLP), CL has been adopted in various contexts ranging from text classification~\cite{wei2019eda}, embedding representation learning~\cite{gao2021simcse} to question answering~\cite{xiong2020approximate, liu2021dense}. CL has also been integrated with PLM as a more effective fine-tuning strategy for downstream tasks \cite{su2021csslm}. In our work, we propose an integration of CL with PLM probing techniques to further refine imperfect PLM-derived segments via (1) unsupervised signals from PLM itself, and (2) less expensive sentence-level intent label supervision for improved SI performance.   
\vspace*{-0.2cm}
\begin{figure}[bt]
    \centering
   \includegraphics[trim={0.0cm 0.2cm 0.0cm 0.0cm},clip, width=\columnwidth]{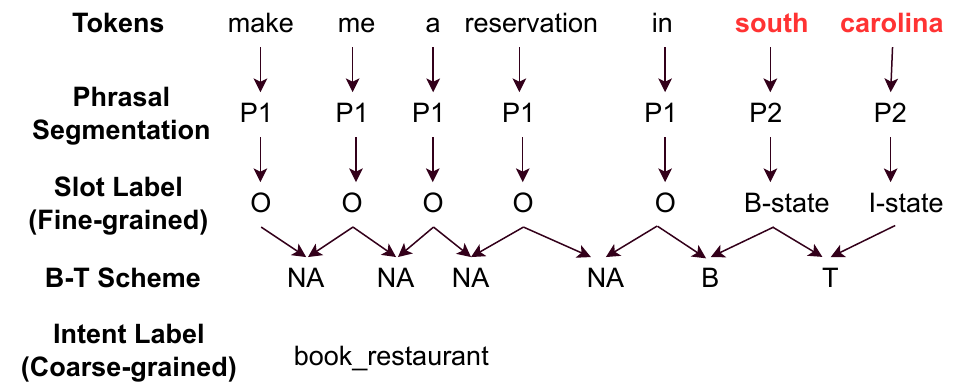} 
    \caption{Illustration of connections between Phrasal Segmentation (PS), Beginning-Inside-Outside (BIO) Tagging Slot Label and Break-Tie (B-T) Labeling Schema based on Golden Slot Labels (\textcolor{red}{Red}: denotes \textcolor{red}{Golden Slot Labels} for the utterance, \textbf{P1,P2} denote identified phrases, \textbf{NA, B,T} denote Not-Relevant, Break, Tie Labels in B-T Labeling Scheme)}
    \label{fig:example}
    \vspace*{-0.6cm}
\end{figure}
\begin{figure*}[bt]
    \centering
   \includegraphics[trim={0.0cm 0.1cm 3.0cm 0.0cm},clip, width=\textwidth]{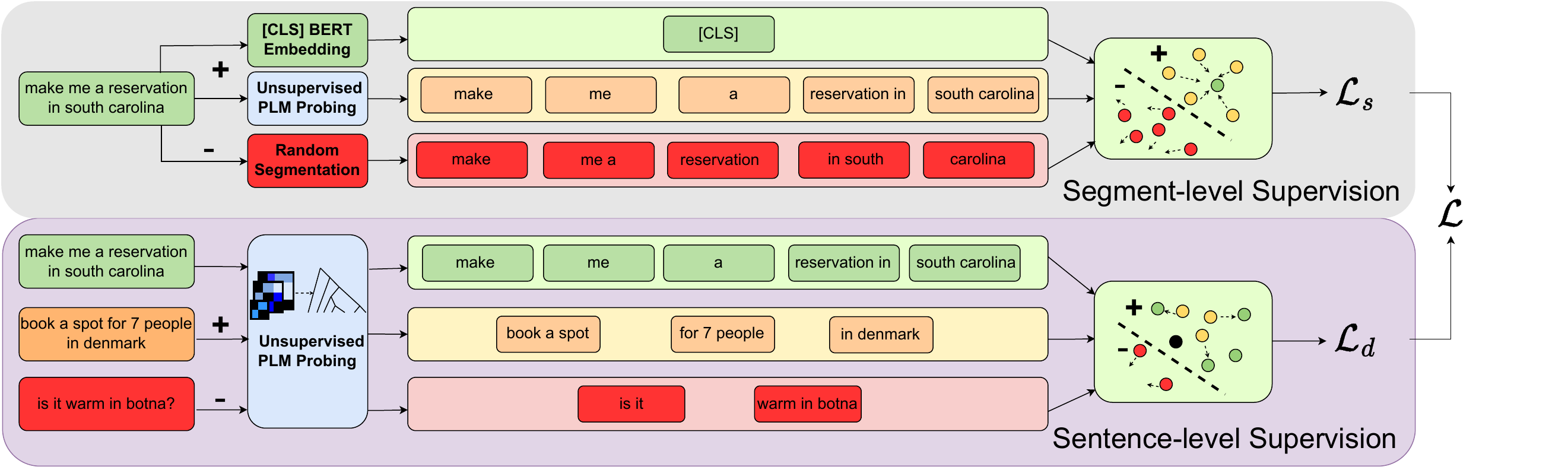}
    \caption{Illustration of the Proposed Model Overview. The model is made up of two-level Contrastive Learning depicted by two modules: (1) \textbf{Segment-level Supervision (SegCL)} via Unsupervised PLM Probing (UPL), (2) \textbf{Sentence-level Supervision (SentCL)}  via intent labels. \textcolor{green}{Green}, \textcolor{orange}{Orange}, \textcolor{red}{Red} denote \textcolor{green}{Anchor}, \textcolor{orange}{Positive}, \textcolor{red}{Negative} samples respectively. \textbf{Black circle} denotes the representation of the \textbf{cropped segment} from Augmentation.}
    \label{fig:overview}
    \vspace*{-0.6cm}
\end{figure*}

\section{Problem Formulation}
\label{sec:prob_form}
\vspace*{-0.2cm}
\paragraph{Slot Induction} We introduce the task of Slot Induction (SI) whose objective is to identify phrases containing token-level slot labels.  
Unlike traditional SF and previously proposed AISI framework \cite{zeng2021automatic}, in our SI task, both slot boundaries and slot types are unknown during training. The task is also related to Phrasal Segmentation/ Tagging (PS) methods \cite{shang2018automated,10.1145/3447548.3467397}. However, there are three key distinctions: (1) utterances and intent labels (if available) are the only sources of information for the task, (2) slot phrases (i.e. close by (\textit{spatial\_relation}), most expensive (\textit{cost\_relative})), are not restricted to noun phrases, (3) slot phrases (i.e. strauss is playing today (\textit{movie\_name})) might be more sophisticated and harder to identify than typical noun phrases (i.e. chicago (\textit{city})). These differences explain why PS methods do not consistently perform well in our proposed SI task (Section \ref{sec:result}).

Specifically, given an utterance with the length of $T$ tokens $x=[x_{1},x_{2}..., x_{T}]$, SI task aims to make decisions at $T-1$ positions whether to (1) tie the current token with the previous one to extend the current phrase \footnote{In our work, we use the term \textbf{segment} and \textbf{phrase} interchangeably.}, or (2) break away from the previous token/ phrase to form a new phrase.
\vspace*{-0.1cm}
\paragraph{Evaluation Metric} We adopt the Break-Tie (B-T) schema \cite{shang2018learning} to evaluate SI task. The metric allows for direct comparison between supervised Sequential Labeling and unsupervised PS methods. In SI setting, \textit{Tie} represents the connection between tokens of the same slot type while \textit{Break} denotes the separation between (1) tokens from different slot types, and (2) tokens from a slot type and non-slot tokens. As the objective of SI is on slot tokens, consecutive non-slot tokens should not contribute to the overall performance. Therefore, additional \textit{NA} labels are introduced to guarantee that evaluations are only conducted on slot tokens and their adjacent tokens.

Figure \ref{fig:example} depicts the connections of SF and PS labels with B-T schema.  For PS, Break denotes the separation of two consecutive phrases. If no phrase is identified by PS methods, every token is considered as \textit{Tie} to one another. In the Figure \ref{fig:example} example, as ``south carolina'' is the only identified phrase, the given sentence is simply split into two phrases where \textit{Break} denotes their junction. Precision, Recall and F-1 Metrics are reported for individual labels, namely B-P,B-R,B-F1 for \textit{Break} and T-P,T-R,T-F1 for \textit{Tie}.

Given an utterance, an optimal SI model makes correct decisions to either break and tie at every token index. Therefore, \textbf{H-Mean}, denoting the harmonic mean between F-1 Scores of \textit{Tie} and \textit{Break} label predictions, is considered the golden criteria for SI model comparison. 
\vspace*{-0.2cm}
\section{Proposed Framework}
\label{sec:framework}
\vspace*{-0.1cm}
In this section, we introduce our proposed Multi-level Contrastive Learning framework for SI task with 2 major components: \textbf{Segment-level Contrastive Learning~(SegCL)} and \textbf{Sentence-level Contrastive Learning (SentCL)} as depicted in Figure \ref{fig:overview}. We first introduce the backbone Unsupervised PLM Probing (UPL) for both components.
\begin{figure}[tb]
    \centering
   \includegraphics[trim={0.0cm 0.0cm 0.0cm 0.2cm},clip, width=\columnwidth]{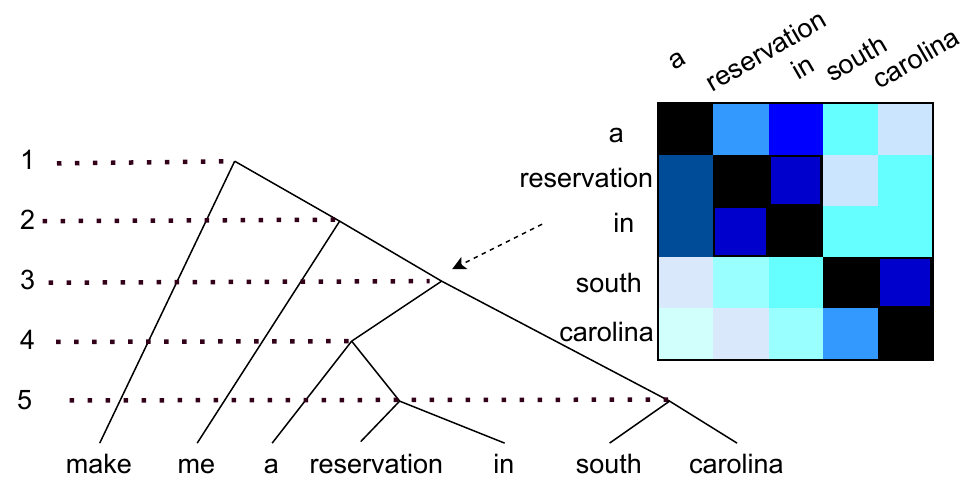}
    \caption{Illustration of UPL Segmentation Tree for sentence \textit{``make me a reservation in south carolina''} with sample Impact Matrix at depth $d=3$ (\textbf{Lighter} color denotes \textbf{lower} impact score). $d=0$ corresponds to the sentence-level representation (no segmentation).}
    \label{fig:pm_example}
    \vspace*{-0.6cm}
\end{figure}
\vspace*{-0.6cm}
\subsection{Unsupervised PLM Probing (UPL)}
\label{subsec:upl}
\vspace*{-0.1cm}
We adopt Token-level Perturbed Masking mechanism~\cite{wu2020perturbed} to construct semantic segments by leveraging PLM in an unsupervised manner. Due to its operations on the output layers of PLM, UPL is flexible with the choices of PLM and avoids local sub-optimal structure from pre-selected PLM layers \cite{clark2019does}. In our study, we use BERT~\cite{devlin2019bert} as an exemplar PLM. Specifically, given a sentence $x = [x_1, \cdots , x_T]$, the Impact Matrix $\mathcal{F} \in \mathbb{R}^{T \times T}$ is constructed by calculating the Impact Score between every possible pair of tokens (including with itself) in the given sentence based on BERT's embedding and a specified distance metric~ \cite{wu2020perturbed}. Leveraging $\mathcal{F}$, UPL derives the structural tree by recursively finding the optimal cut position $k$ with the following objective:
\vspace*{-0.3cm}
\begin{equation}
\vspace*{-0.1cm}
    \label{eq:cut_score}
    \begin{aligned}
    \underset{k}{arg max} (\mathcal{F}_{i..k}^{i..k} + \mathcal{F}_{k+1..j}^{k+1..j} \\
                        - \mathcal{F}_{i..k}^{k+1..j} - \mathcal{F}_{k+1..j}^{i..k})
        \end{aligned}
        \vspace*{-0.1cm}
\end{equation}      
 where $i,j \in [0,T-1]$ denotes the start and end indexes of the segment considered for splitting.  

At every tree depth, sets of combined tokens are considered semantic segments since they preserve certain meanings within utterances. Segments at a deeper level include (1) all segments obtained from previous levels and (2) new segments obtained at the current level. For instance, at depth $d=3$ of the given example in Figure \ref{fig:pm_example}, the obtained segments are \textit{``make''}, \textit{``me''}, \textit{``a reservation in''}, \textit{``south carolina''}. As PLM parameters are updated during training, the derived UPL trees from the same utterance can vastly change. For simplicity, we set the tree depth $d$ as a tunable hyperparameter. 


Formally, at a specified depth $d$ with $m$ semantic segments acquired from UPL, the final representation of the input sentence $x$ is defined as follows:
\vspace*{-0.4cm}
\begin{equation}
\mathbf{h_{U}} = [\overrightarrow{s_{0}}, ... \overrightarrow{s_{m-1}}],  \; \overrightarrow{s_{i}}=\frac{\sum_{j=c}^{d}{\overrightarrow{h_{j}}}}{d-c+1}
    \label{eq:plm}
    \vspace*{-0.2cm}
\end{equation}
where $\bf{h_{U}} \in \mathbb{R}^{m \times d_{h}}$, $d_{h}$ is hidden dimensions of BERT representations, $c$,$d$ are the start and end indexes of the corresponding segment $s_{i}$ and $\overrightarrow{h_{j}}$ represents the BERT embedding of $j$-th token.
\vspace*{-0.2cm}
\subsection{Multi-level Contrastive Learning}
\vspace*{-0.1cm}
As UPL only considers token interactions for segment formation, its semantic segments are far from perfect.  Additional refinements are needed to enhance the quality of the extracted segments via (1) semantic signals captured in segment-level PLM representations, (2) sentence-level intent labels. 

Our overall learning objective is summarized as $ \mathcal{L} = \delta \mathcal{L}_{s} + \gamma \mathcal{L}_{d}$, where $ \mathcal{L}_{s},  \mathcal{L}_{d}$ denote SegCL Loss and SentCL Loss, and $\gamma, \delta$ are their corresponding loss coefficient hyperparameters for aggregation. 
For each CL level, positive and negative samples are drawn separately based on (1) the same batch of sampled anchor samples, (2) different selection criteria detailed below. 


\paragraph{Segment-level Contrastive Learning (SegCL)} UPL produces semantic segments by purely considering the exhaustive word-pair interactions within given sentences. However, it does not take into consideration the overall semantic representation produced by the PLM BERT via special [CLS] tokens. 
Therefore, we propose leveraging [CLS] representations to guide UPL towards more discriminative segment representations via SegCL objectives. Specifically, SegCL aims to minimize the distance between [CLS] representation and UPL segment representations while maximizing the distance between representations of [CLS] and random segments of the corresponding utterance. 

Given a sample utterance, segment representation obtained from UPL is considered a positive sample while negative samples are represented as  segments produced by randomly chosen indexes within the given utterance. The number of segments for both positive and negative samples are kept similar ($m$) so that SegCL focuses on learning the optimal locations of segmentation indexes. We adopt InfoNCE contrastive loss~\cite{oord2018representation}:
\vspace*{-0.3cm}
\begin{equation}
             \mathcal{L}_{s} =-log \frac{\exp^{cos(\overrightarrow{h_{C}}, \bf{h_{U}})/ \tau_{s}}}
                    {\exp^{cos(\overrightarrow{h_{C}}, \bf{h_{U}})/ \tau_{s}} + \exp^{cos(\overrightarrow{h_{C}}, \bf{h_{r}})/ \tau_{s}}}
    \vspace*{-0.1cm}
\end{equation}
where $\overrightarrow{h_{C}} \in \mathbb{R}^{1 \times {d_{h}}}$ denotes [CLS] representation from BERT, and $\bf{h_{U}}, \bf{h_{r}} \in \mathbb{R}^{m \times d_{h}}$ denote the representations from UPL and random segmentation. $m$ is the number of extracted segments from UPL as defined in Equation \ref{eq:plm}. $\tau_{s}$ is the soft segment-level temperature hyperparameter. 

\paragraph{Sentence-level Contrastive Learning (SentCL)}
 Besides relying on UPL, we propose leveraging sentence-level intent labels to further improve the quality of segment representations derived from UPL. Specifically, we randomly draw positive and negative samples based on the intent labels of the given anchor samples. As utterances with similar intents tend to share common slot phrases, our SentCL aims to learn discriminative segments for better alignment between utterances from the same intents. We adopt InfoNCE loss for SentCL:
\vspace*{-0.2cm}
\begin{equation}
     \mathcal{L}_{d} =-log \frac{\exp^{cos(\bf{h_{a}}, \bf{h_{+}})/ \tau_{d}}}
                    {\exp^{cos(\bf{h_{a}}, \bf{h_{+}})/ \tau_{d}} + \exp^{cos(\bf{h_{a}}, \bf{h_{-}})/ \tau_{d}}}     
\end{equation}
where $\bf{h_{a}} \in \mathbb{R}^{m \times d_{h}}, \bf{h_{+}} \in \mathbb{R}^ {a \times d_{h}}, \bf{h_{-}} \in \mathbb{R}^{b \times d_{h}}$ denote the representations of anchor, positive and negative samples respectively and $m,a,b$ denote the number of extracted segments from UPL for the respective samples. $\tau_{d}$ is the soft sentence-level temperature hyperparameter. 

To further encourage the model to identify discriminative segments from the same sentence-level intent label, we adopt random segment cropping as an augmentation strategy. As UPL could generate a vastly different number of segmentation based on the the cut\_score (Equation \ref{eq:cut_score}) from the updated BERT parameters at each step, we conduct random segmentation cropping by a percent ratio ($\beta$) so that it could be adapted to individual input utterances and segmentation trees. The remaining segments after cropping are utilized to compute $\mathcal{L}_{d}$. 

\section{Experiments}
\vspace*{-0.2cm}
\subsection{Datasets \& Evaluation Tasks}
\vspace*{-0.1cm}
We evaluate our proposed work on the two publicly available NLU benchmark datasets ATIS~\cite{tur2010left} and SNIPS~\cite{coucke2018snips} with the previously proposed data splits~\cite{zhang2019joint}. 

To evaluate the generalization of the refined representations from our proposed work, we conduct additional splits of each dataset into 2 parts (P1 and P2). For each benchmark dataset, we construct P1 for SI evaluation by reserving samples from randomly chosen 60\% of available intents. The remaining samples (P2) are used as test sets for evaluating SF task when generalized towards emerging intents. The objective of this splitting strategy is two-fold: (1)  Since there is no overlapping intent between P1 and P2, there exists no information leakage of intents leveraged in SI training (P1) while evaluating SF (P2). (2) We can validate the generalization capability of representations learned from our SI framework in other slot-related tasks. Statistics for both parts of each dataset are reported in Table \ref{detailsdataset}.

\begin{table}[tb]
\centering
\caption{Details of SNIPS and ATIS datasets.}
 \vspace*{-0.2cm}
\resizebox{\columnwidth}{!}{%
\begin{tabular}{|c|c|c|c|c|}
\hline 
& \textbf{SNIPS\_P1} & \textbf{SNIPS\_P2} & \textbf{ATIS\_P1} & \textbf{ATIS\_P2} \\
\hline 
\# Intents & 5 & 2 & 14 & 7\\
\# Slots & 31 & 16 & 68 & 63 \\
\# Train Samples & 9356 & -- & 3811 &-- \\
\# Validation Samples & 500 & -- & 414 & -- \\
\# Test Samples & 501 & 4127 & 750 & 895 \\
Avg Train Sent Length & 8.65  &  -- & 11.67 & -- \\
Avg Valid Sent Length  & 8.72 & -- & 11.82 & --\\
Avg Test Sent Length  & 8.71 & 9.87 & 10.68 & 8.92 \\
\hline
\end{tabular}%
}
 \vspace*{-0.6cm}
\label{detailsdataset}
\end{table}

\begin{table*}[htb]
\centering
\caption{Experimental performance result on SNIPS dataset over 3 runs (\textbf{H-Mean} is considered the golden criteria for SI (Section \ref{sec:prob_form})). $^{\P}$ denotes models that do not require random initializations.}
\vspace*{-0.2cm}
\resizebox{\textwidth}{!}{%
\begin{tabular}{|c||c|c||c|c|c||c|c|c||c||}
\hline 
 & \textbf{Model} & \textbf{Prior Knowledge} & \multicolumn{3}{c||}{Break} & \multicolumn{3}{c||}{Tie} & \multicolumn{1}{c||}{\textbf{H-Mean}}\\
\specialrule{.1em}{0.05em}{.05em}
 & & & B-P & B-R & B-F1 & T-P & T-R & T-F1 &   \\
\specialrule{.1em}{0.05em}{.05em}

\multirow{3}{*}{\textbf{Upper Bound}} & Joint BERT FT & Slot + Intent & 96.91 $\pm$ 0.17 & 96.62 $\pm$ 0.69 & 96.76 $\pm$ 0.26 & 73.55 $\pm$ 0.38 & 73.39 $\pm$ 1.03 & 73.47 $\pm$ 0.38 & 83.52 $\pm$ 0.16 \\
& FlairNLP $^{\P}$ & POS \& NER &80.04 & 62.81 & 70.38 & 48.25 & 63.31 & 54.77 & 61.60 \\
 & SpaCy $^{\P}$ & POS & 75.73 & 50.29 & 60.45 & 41.71 & 62.97 & 50.18 & 54.84 \\

\hline 
\multirow{5}{*}{\textbf{Comparable}} & DP-LB $^{\P}$ & -- & 59.68 & 34.27 & 43.54 & 21.69 & 38.53 & 27.76 & 33.90\\
 & DP-RB $^{\P}$ & -- & 66.53 & 52.56 & 58.73 & 33.97 & 52.24 & 41.17 & 48.40 \\

& AutoPhrase & External KB & 65.51 $\pm$ 0.23 & 57.16 $\pm$ 2.59 & 61.05 $\pm$ 1.15 & 33.39 $\pm$ 0.74 & 36.62 $\pm$ 1.67 & 34.93 $\pm$ 1.50 & 44.43 $\pm$ 1.64 \\
        
& UCPhrase & PLM & 42.25 $\pm$ 4.90 & 20.26 $\pm$ 2.71 & 27.39 $\pm$ 1.95 & 36.06 $\pm$ 2.42 & \textbf{73.53 $\pm$ 3.33} & \textbf{48.39 $\pm$ 2.91} & 34.98 $\pm$ 2.35 \\

& USSI $^{\P}$ & PLM  & \textbf{83.21} & 62.12 & 71.14 & 33.96 & 49.93 & 40.42 & 51.55 \\
  
 \hline

& Ours (w/o CL) $^{\P}$ & PLM & 75.36 & 66.70  & 70.76  & 38.51 & 45.81 & 41.84 & 52.59 \\

 & Ours (w/o SentCL) & PLM  & 76.09 $\pm$ 0.73 & 66.43 $\pm$ 0.29 & 70.94 $\pm$ 0.49 & 39.15 $\pm$ 0.60 & 47.9 $\pm$ 0.91 & 43.09 $\pm$ 0.73 & 53.61 $\pm$ 0.71 \\  
 & \textbf{Ours (full)} & \textbf{PLM + Intent}  & 76.87 $\pm$ 0.25 & \textbf{67.77 $\pm$ 0.26} & \textbf{72.00 $\pm$ 0.24} & \textbf{40.39 $\pm$ 0.16} & 48.49 $\pm$ 0.19 & 44.07 $\pm$ 0.04 & \textbf{54.68 $\pm$ 0.08} \\ 
\hline
\end{tabular}%
}
\label{snipsresult}
\vspace*{-0.1cm}
\end{table*}

\begin{table*}[htb]
\centering
\caption{Experimental performance result on ATIS dataset over 3 runs (\textbf{H-Mean} is considered the golden criteria for SI (Section \ref{sec:prob_form})). $^{\P}$ denotes models that do not require random initializations.}
\vspace*{-0.2cm}
\resizebox{\textwidth}{!}{%
\begin{tabular}{|c||c|c||c|c|c||c|c|c||c||}
\hline 
 & \textbf{Model} & \textbf{Prior Knowledge} & \multicolumn{3}{c||}{Break} & \multicolumn{3}{c||}{Tie} & \multicolumn{1}{c||}{\textbf{H-Mean}}\\
\specialrule{.1em}{0.05em}{.05em}
 &  &  & B-P & B-R & B-F1 & T-P & T-R & T-F1 &   \\
\specialrule{.1em}{0.05em}{.05em}

\multirow{3}{*}{\textbf{Upper Bound}}  & Joint BERT FT & Slot + Intent & 98.49 $\pm$ 0.24  & 99.33 $\pm$ 0.08 & 98.91 $\pm$ 0.09& 59.07 $\pm$ 0.36 & 58.27 $\pm$ 0.89 & 58.67 $\pm$ 0.63 & 73.65 $\pm$ 0.54 \\

& FlairNLP $^{\P}$ & POS \& NER  & 95.44 & 77.90 & 85.78 & 41.34 & 61.91 & 49.58 & 62.84 \\
 & SpaCy $^{\P}$  & POS & 94.45 & 69.64 & 80.17 & 35.33 & 61.17 & 44.79 & 57.47 \\
\hline 
\multirow{5}{*}{\textbf{Comparable}} &  DP-LB $^{\P}$ & --  & 80.80 & 36.38 & 50.17 & 12.32 & 38.51 & 18.67 & 27.21 \\
 & DP-RB $^{\P}$ &  -- & 84.24 & \textbf{66.84} & \textbf{74.54} & 14.81 & 30.52 & 19.94 & 31.46 \\

 & AutoPhrase & External KB & 75.96 $\pm$ 0.04 & 40.06 $\pm$ 0.28 & 52.46 $\pm$ 0.18 & 19.75 $\pm$ 0.21 & 49.33 $\pm$ 0.38 & \textbf{28.20 $\pm$ 0.28} & 36.68 $\pm$ 0.21 \\
        
 & UCPhrase & PLM  & 47.25 $\pm$ 0.04 & 17.27 $\pm$ 0.72 & 25.29 $\pm$ 0.78 & 17.36 $\pm$ 0.16 & \textbf{58.21 $\pm$ 0.68} & 26.75 $\pm$ 0.11 & 26.00 $\pm$ 0.47 \\

  & USSI $^{\P}$ & PLM  & \textbf{95.06} & 56.36 & 70.77 & 14.78 & 45.22 & 22.28 & 33.89 \\
 
 \hline

  & Ours (w/o CL) $^{\P}$ & PLM & 86.40  & 61.53 & 71.87  & 18.23  & 35.27 & 24.04  & 36.03  \\
  
 & Ours (w/o SentCL) & PLM & 87.29 $\pm$ 0.15 & 64.21 $\pm$ 0.27 & 73.99 $\pm$ 0.13 & 20.09 $\pm$ 0.08 & 35.86 $\pm$ 0.35 & 25.75 $\pm$ 0.08 & 38.20 $\pm$ 0.08 \\
 
 & \textbf{Ours (full)} & \textbf{PLM + Intent} & 87.80 $\pm$ 0.27 & 63.27 $\pm$ 0.67 & 73.54 $\pm$ 0.36 & \textbf{20.53 $\pm$ 0.14} & 37.89 $\pm$ 0.99 & 26.63 $\pm$ 0.26 & \textbf{39.10 $\pm$ 0.24}  \\
\hline
\end{tabular}%
}
\label{atisresult}
\vspace*{-0.6cm}
\end{table*}
\vspace*{-0.1cm}
\paragraph{\underline{Evaluation Task 1:} \textbf{Slot Induction (P1)}} We conduct evaluation of Unsupervised SI task on P1 of both SNIPS and ATIS datasets. B-T evaluation metrics are adopted as introduced in Section \ref{sec:prob_form}. Implementation details of our SI model, including hyperparameters, are discussed in Appendix \ref{sec:ap_hyper}.
\paragraph{\underline{Evaluation Task 2:} \textbf{Generalization towards Emerging Intents (P2)}} To evaluate the generalization of SI refinement, we conduct SF training on P1 datasets with different BERT initializations (Original vs Refined BERT) and evaluation on emerging intents and slots in P2. Slot Precision (S-P), Recall (S-R), F1 (S-F1) are reported on P2. Implementation is detailed in Appendix \ref{sec:ap_sf}.


\subsection{Slot Induction Baseline}
\vspace*{-0.1cm}
We conduct a comprehensive study that evaluates our SI approach with both \textit{Upper Bound} and \textit{Comparable} Methods. For fair comparisons, we leverage the same ``bert-base-uncased" PLM \cite{devlin2019bert} across all applicable baselines. The \textit{Upper Bound} includes methods that leverage directly \textbf{token-level labels} such as Golden Slot Labels, Named Entity Recognition (NER) Labels, Part-of-Speech (POS) Tagging or Noun Phrase (NP) Labels during training and/or pre-training process, including \textbf{Joint BERT FT}, \textbf{SpaCy}~\cite{Honnibal_spaCy_Industrial-strength_Natural_2020}, \textbf{FlairNLP}~\cite{akbik2018coling}.

In addition, we compare with other \textbf{unsupervised} PS methods that do not require any token-level labels as \textit{Comparable} Baselines, including: \textbf{Dependency Parsing (DP-RB/DP-LB)}, \textbf{AutoPhrase}~\cite{shang2018automated}, \textbf{UCPhrase}~\cite{10.1145/3447548.3467397}, \textbf{USSI}~\cite{yu2022unsupervised}. For fair comparisons with \textit{Comparable} baselines, we also report results from our model's variants with similar prior knowledge assumption, namely \textbf{Ours (w/o CL)}, \textbf{Ours (w/o SentCL)} . Due to space constraints, details of \textit{Upper Bound} and \textit{Comparable} baselines are provided in Appendix \ref{sec:ap_base_up}, \ref{sec:ap_base_comp} respectively. 


\begin{figure*}[bt]
    \centering
    \subfloat[\centering Segment-level Supervised Positive-Anchor Pair ]{{\includegraphics[trim={0.4cm 6.0cm 12.0cm 4.8cm},clip, width=0.85\columnwidth]{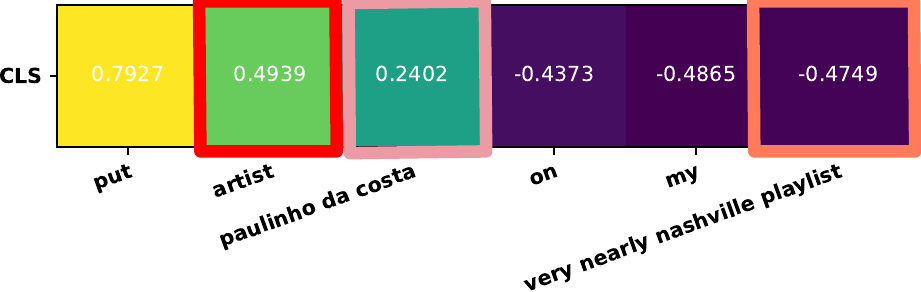} } \label{subfig:self_pos_anchor}}%
    \qquad
    \subfloat[\centering Segment-level Supervised Negative-Anchor Pair]{{\includegraphics[trim={0.4cm 6.0cm 12.0cm 4.8cm},clip, width=0.85\columnwidth]{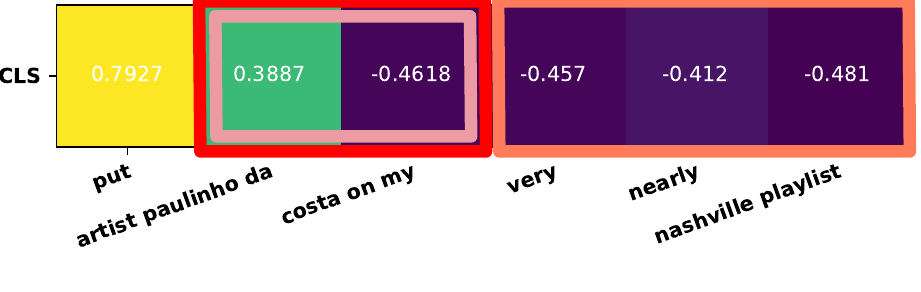}}
    \label{subfig:self_neg_anchor}}
      \qquad
        \subfloat[\centering Sentence-level Supervised Positive-Anchor Pair  ]{{\includegraphics[trim={0.5cm 4.0cm 13.5cm 0.9cm},clip, width=0.85\columnwidth]{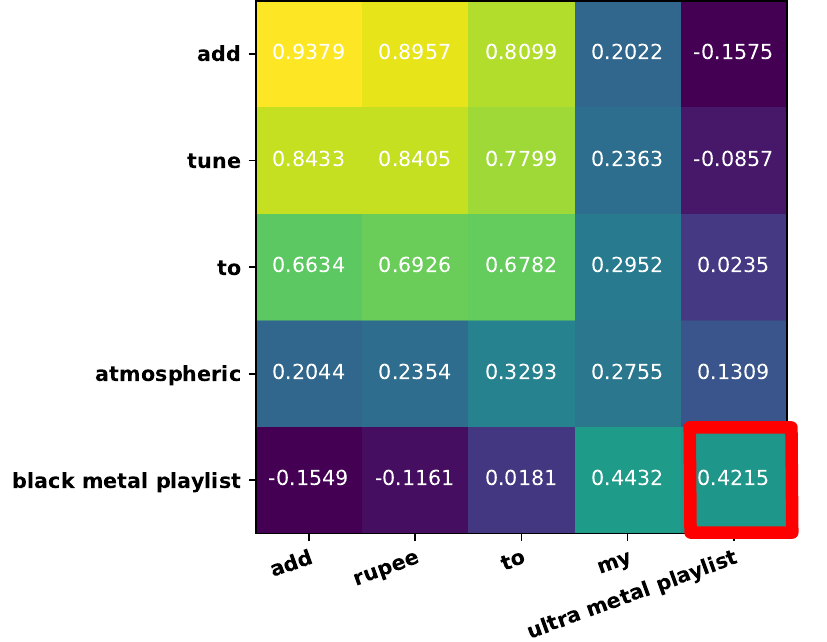}}
        \label{subfig:dist_pos_anchor}}%
    \qquad
    \subfloat[\centering Sentence-level Supervised Negative-Anchor Pair ]{{\includegraphics[trim={0.4cm 4.0cm 13.0cm 0.9cm},clip, width=0.85\columnwidth]{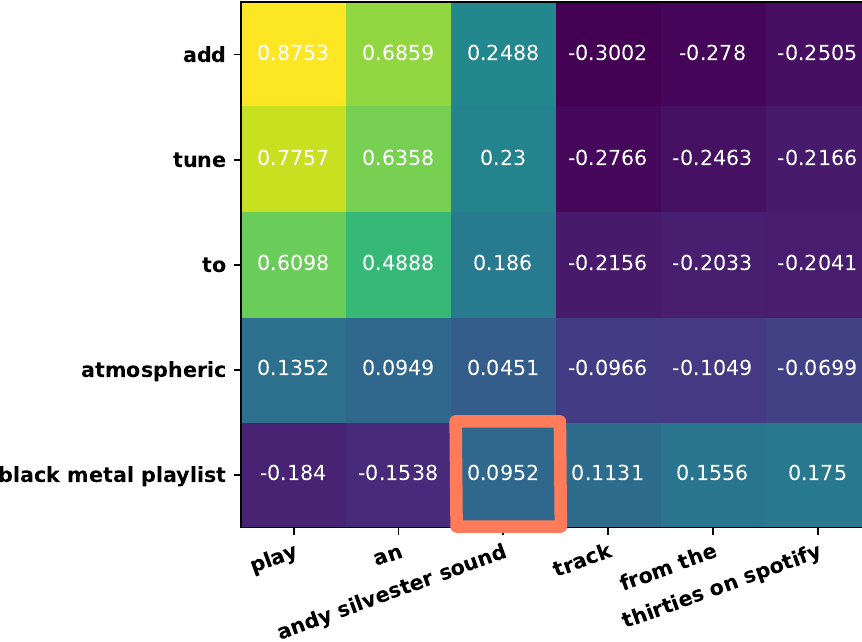}}
    \label{subfig:dist_neg_anchor}}%
    \vspace*{-0.2cm}
    \caption{Similarity Matrices between positive/negative and anchor samples from SegCL and SentCL. For SegCL (\protect\subref{subfig:self_pos_anchor}, \protect\subref{subfig:self_neg_anchor}), positive-anchor pair is more aligned as the sum of similarity scores between positive segments and [CLS] representation (i.e. sum of row-wise cell values) is higher than the negative counterpart. Boundaries of all slot types (presented by \textcolor{red}{red}, \textcolor{pink}{pink}, \textcolor{orange}{orange} boxes) are correctly recognized in the positive sample in contrast to the negative counterpart. For SentCL (\protect\subref{subfig:dist_pos_anchor}, \protect\subref{subfig:dist_neg_anchor}), positive-anchor pair assigns a higher similarity score to the aligned slot phrase (\textcolor{red}{red box}) while negative-anchor pair reduces similarity scores between potential relevant slot phrase (\textcolor{orange}{orange box}).} %
    \label{fig:attn_case}
    \vspace*{-0.6cm}
\end{figure*}

\section{Result \& Discussion}
\label{sec:result}
\vspace*{-0.2cm}
\subsection{Slot Induction}
\vspace*{-0.1cm}
From our experimental results in Table \ref{snipsresult} and \ref{atisresult}, for SI task, our proposed framework outperforms the \textit{Comparable} Methods in H-Mean evaluation metric for B-T schema on both datasets. We achieve significant gains in SNIPS dataset (+6.28 points in H-Mean as compared to the next \textit{Comparable} Methods). Despite lack of access to any types of token-level labels, our method is also closely on par with some of the \textit{Upper Bound} methods that have been pre-trained with token-level labels (0.16 point difference from SpaCy in H-Mean). Despite promising achievements, most unsupervised PS methods only achieve competitive Break performance as compared to supervised methods but fall behind more significantly in terms of Tie performance. This implies unsupervised methods are able to differentiate non-slot tokens from slot tokens but tend to fragment slot tokens of the same type into multiple slot phrases due to the missing knowledge of token-level slot label spans. 

UCPhrase is an exceptional baseline as it achieves significant better Tie but worse Break performance as compared to other \textit{Comparable} baselines. This roots from the lack of keyphrases predicted from the model, leading to higher tendency to ``tie'' tokens. We speculate that its core phrase miner's dependency on frequency is not effective for extracting slots in NLU tasks. Phrases with high frequency in utterances are typically non-slot tokens (i.e. add, reserve), leading to limited meaningful core phrases for phrase-tagging training. 

\newcommand*{\indentation}{\hspace*{0.2cm}}%
\begin{table}[bt]
\centering
\caption{Ablation study of effectiveness of SegCL and SentCL on SNIPS and ATIS in terms of H-Mean}
\vspace*{-0.3cm}
\resizebox{0.9\columnwidth}{!}{%
\begin{tabular}{|c||c|c|c|c||}
\hline 
  & \multicolumn{1}{c|}{\textbf{SNIPS}} & \multicolumn{1}{c|}{\textbf{ATIS}} \\
 \hline

Ours (w/o CL) & 52.59 & 36.03 \\
\hline
\indentation + 
SegCL
& 53.61 $\pm$ 0.71 & 38.20 $\pm$ 0.08 \\
\indentation + 
SentCL
(w/o aug) & 53.44 $\pm$ 0.22 & 37.59 $\pm$ 0.81 \\
\indentation + 
SentCL (w aug) & 54.23 $\pm$ 0.10 & 38.12 $\pm$ 0.36 \\
\hline
\textbf{Ours (full)} & \textbf{54.68 $\pm$ 0.08}& \textbf{39.10 $\pm$ 0.24} \\
\hline
\end{tabular}%
}
 \vspace*{-0.7cm}
\label{tab:ablation}
\end{table}
On ATIS dataset, the gap between \textit{Comparable}  Methods and \textit{Upper Bound} is more significant as utterances tend to be longer and contain a wider variety of slot types than SNIPS dataset. This leads to a significant reduction in T-P across all of the \textit{Comparable} Methods, resulting in a larger gap in H-Mean for ATIS dataset (approximately 18.37 points in comparison with 0.16 points in SNIPS dataset). Additionally, in comparison with SNIPS dataset, ATIS dataset contains more domain-independent slot types such as \textit{city\_name} (New York), \textit{country\_name} (United States). Therefore, methods leveraging either relevant token-level labels (i.e. POS, NER tags) or additional large-scaled external Knowledge Base (i.e. Wikipedia) achieve considerable performance gains. For instance, \textit{FlairNLP} is only 10.81 points below the Fully Supervised \textit{Joint BERT FT} on ATIS dataset (as compared to 21.92 points below on SNIPS) in terms of H-Mean.

Compared with USSI, \textit{Ours (w/o CL)} consistently achieves better H-Mean performance on both ATIS and SNIPS datasets (1.04\% and 2.14\% respectively). We hypothesize USSI might suffer from the local sub-optimality of pre-selected layers within deep PLM architecture. As the attention distribution across different layers varies \cite{clark2019does}, the pre-selected layers can significantly impact the unsupervised semantic probing of PLM.  

Table \ref{tab:ablation} demonstrates that both SegCL and SentCL (w aug) objectives provide valuable information for SI task, leading to improved performance on both datasets beyond \textit{Ours (w/o CL)}. 
\vspace*{-0.2cm}
\paragraph{Segment-level Supervision (SegCL)} As observed in Figure \ref{subfig:self_pos_anchor}, \ref{subfig:self_neg_anchor}, semantic representation of the given utterance via [CLS] token is closer to the UPL-derived segments as compared to random segment counterparts due to the higher sum of similarity score (0.1281 > -0.6304). UPL segments also correctly identify nearly all of the slot ground truth labels ~(i.e. artist (\textit{music\_item}), paulinho da costa (\textit{artist}), my (\textit{playlist\_owner}), very nearly nashville (\textit{playlist})) in the given utterance while random segmentations truncate the slot phrases incorrectly. 

\begin{figure}[bth]
    \centering
    \includegraphics[trim={1.1cm, 0.5cm, 9.5cm, 3.5cm},clip,width=0.95\columnwidth]{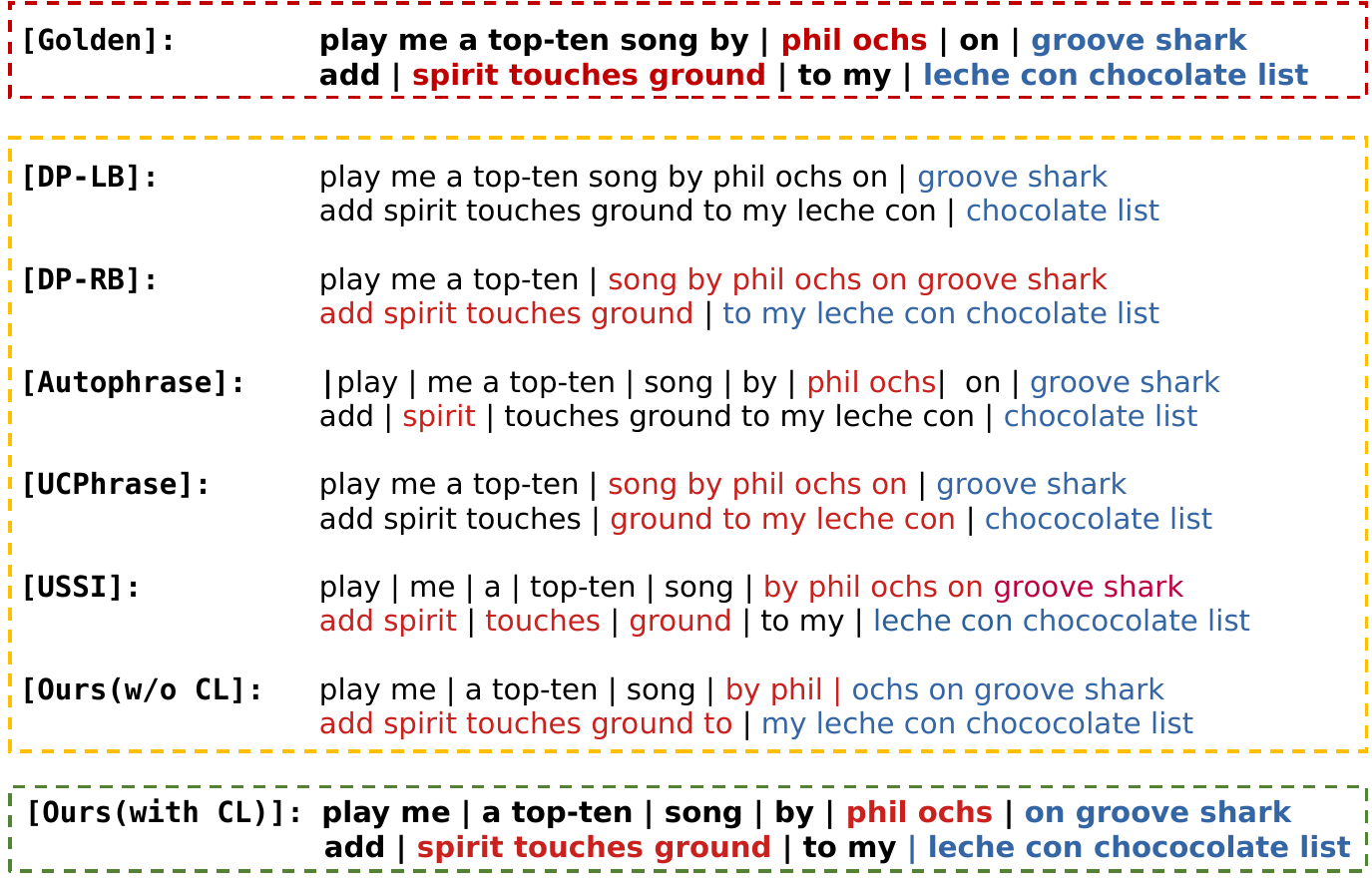}
    \vspace*{-0.2cm}
    \caption{Sample Segmentation Results from \textit{Comparable} Methods in comparison with \textbf{Golden Slot Labels} on SNIPS dataset where ``|'' denotes the \textit{Break} as introduced in Figure \ref{fig:example}. \textcolor{red}{Red}, \textcolor{blue}{Blue} denote distinct slot label segments. The colors are repeated in \textit{Comparable} Methods to showcase the consistency of models' predictions with ground truth labels under the condition no more than 2 tokens in the segments are mispredicted.}
    \label{fig:case}
     \vspace*{-0.6cm}
\end{figure}

\begin{figure*}[tb]
    \centering
     \subfloat[\centering Pre-trained BERT (Train Slots) ]{{\includegraphics[trim={1.8cm 1.4cm 3.5cm 2.9cm},clip, width=0.85\columnwidth]{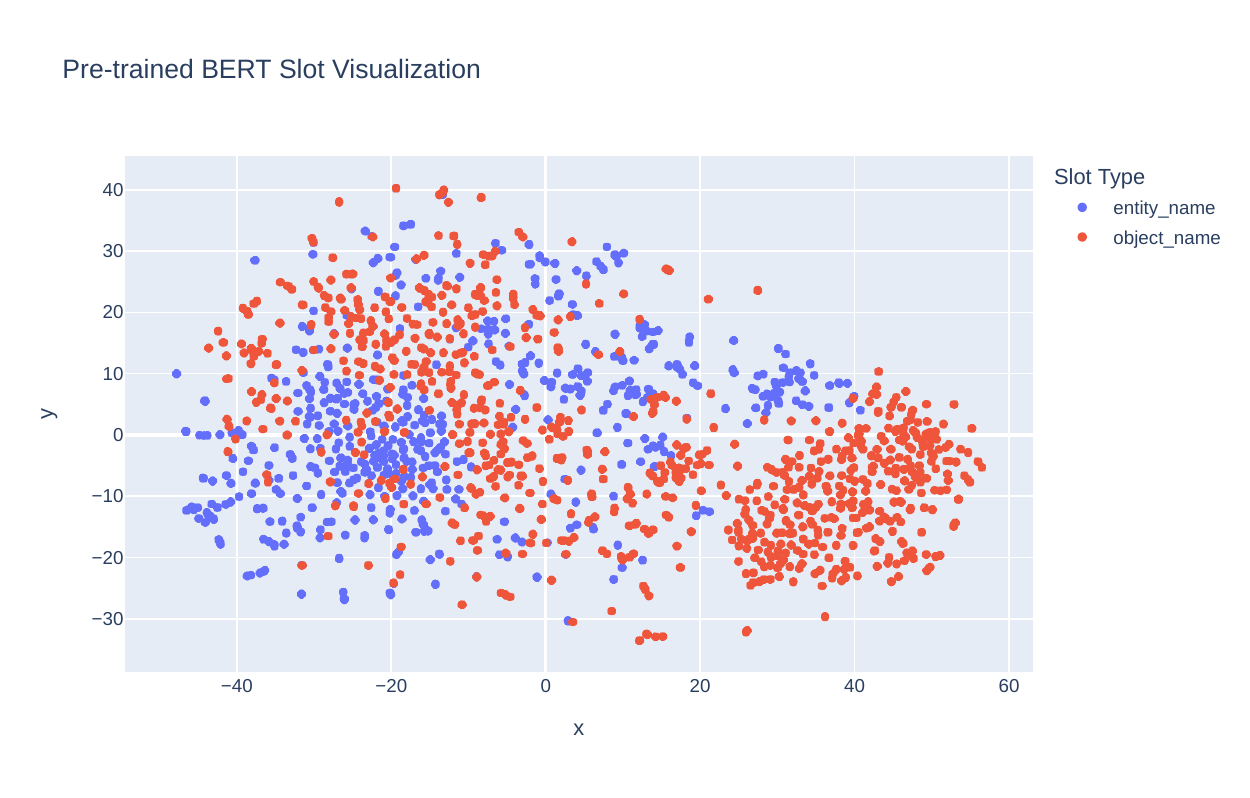} }
      \label{subfig:pre_train}}%
     \qquad
     \subfloat[\centering Refined BERT (Train Slots)]{{\includegraphics[trim={1.8cm 1.4cm 3.5cm 2.9cm},clip, width=0.85\columnwidth]{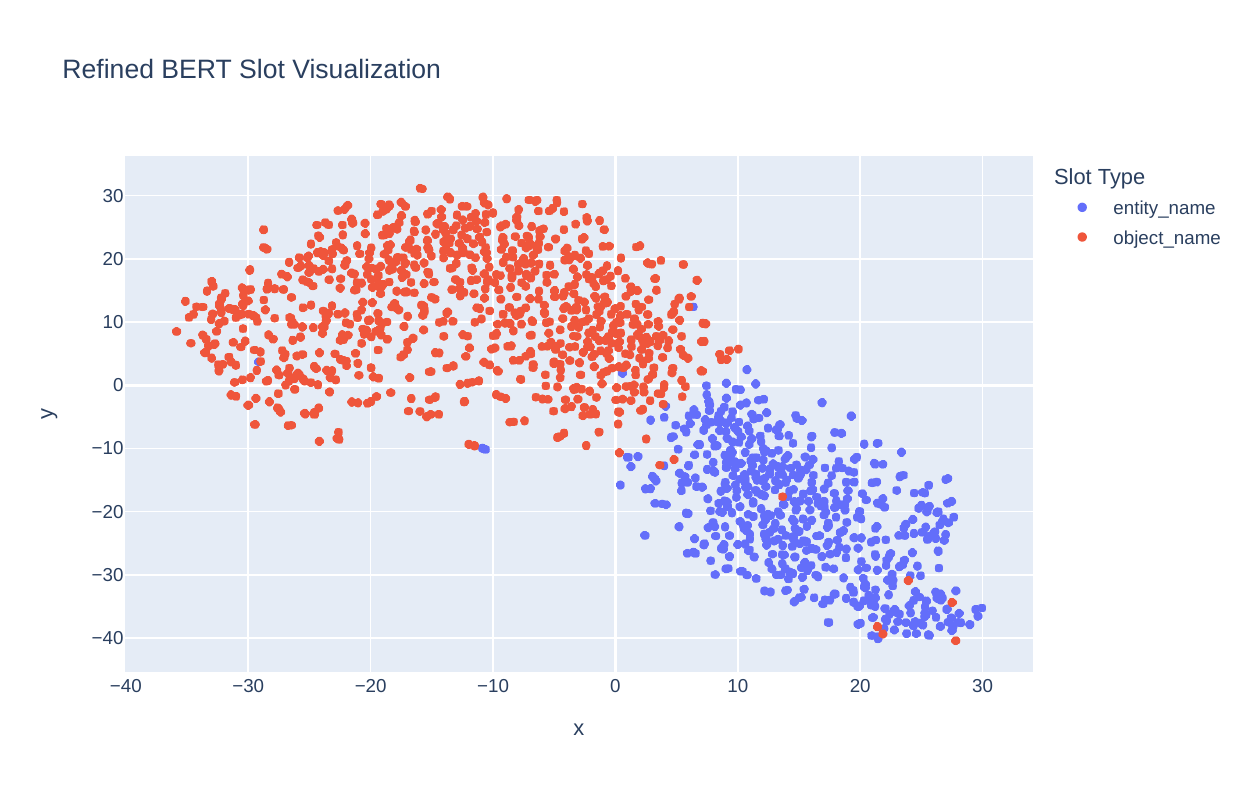} }
     \label{subfig:refined_train}
     }%
     \qquad
        \subfloat[\centering Pre-trained BERT (Test Slots) ]{{\includegraphics[trim={1.8cm 1.4cm 3.5cm 2.9cm},clip, width=0.85\columnwidth]{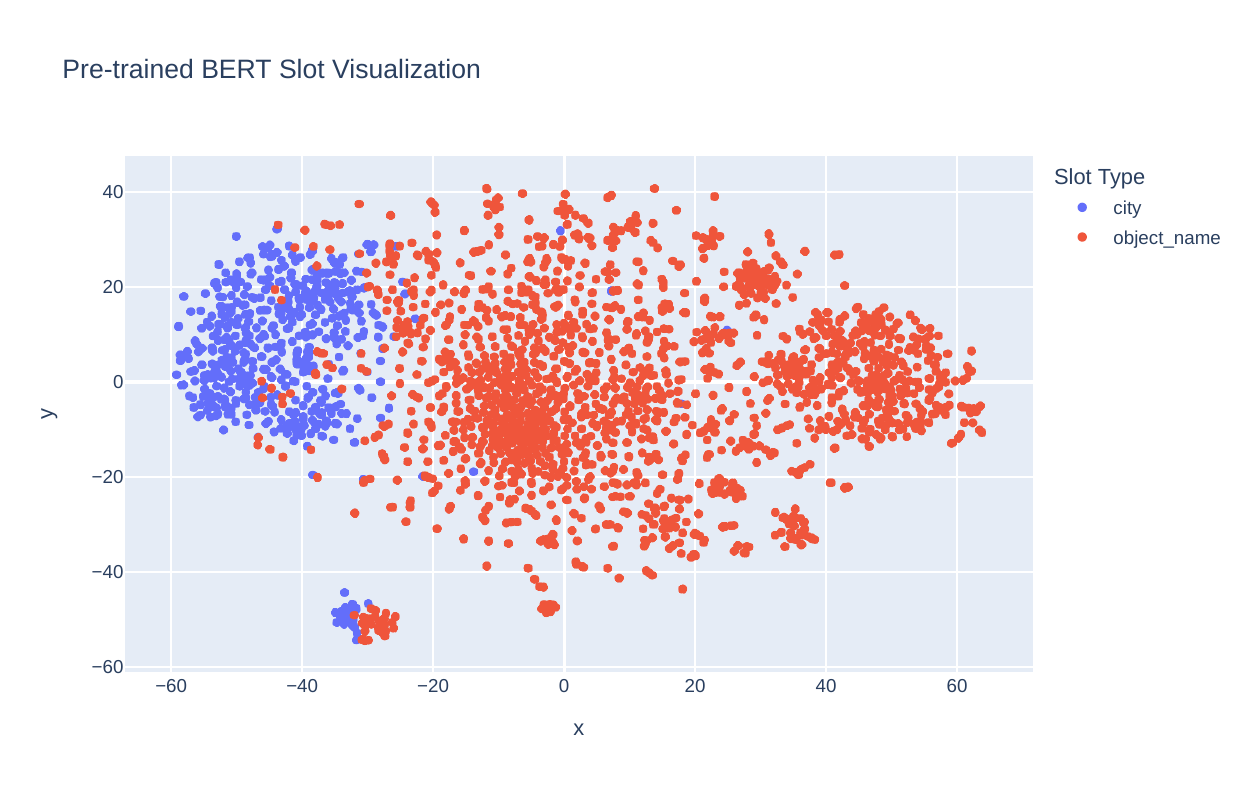}}
        \label{subfig:pre_test}
        }%
    \qquad
    \subfloat[\centering Refined BERT (Test Slots)]{{\includegraphics[trim={1.8cm 1.4cm 3.5cm 2.9cm},clip, width=0.80\columnwidth]{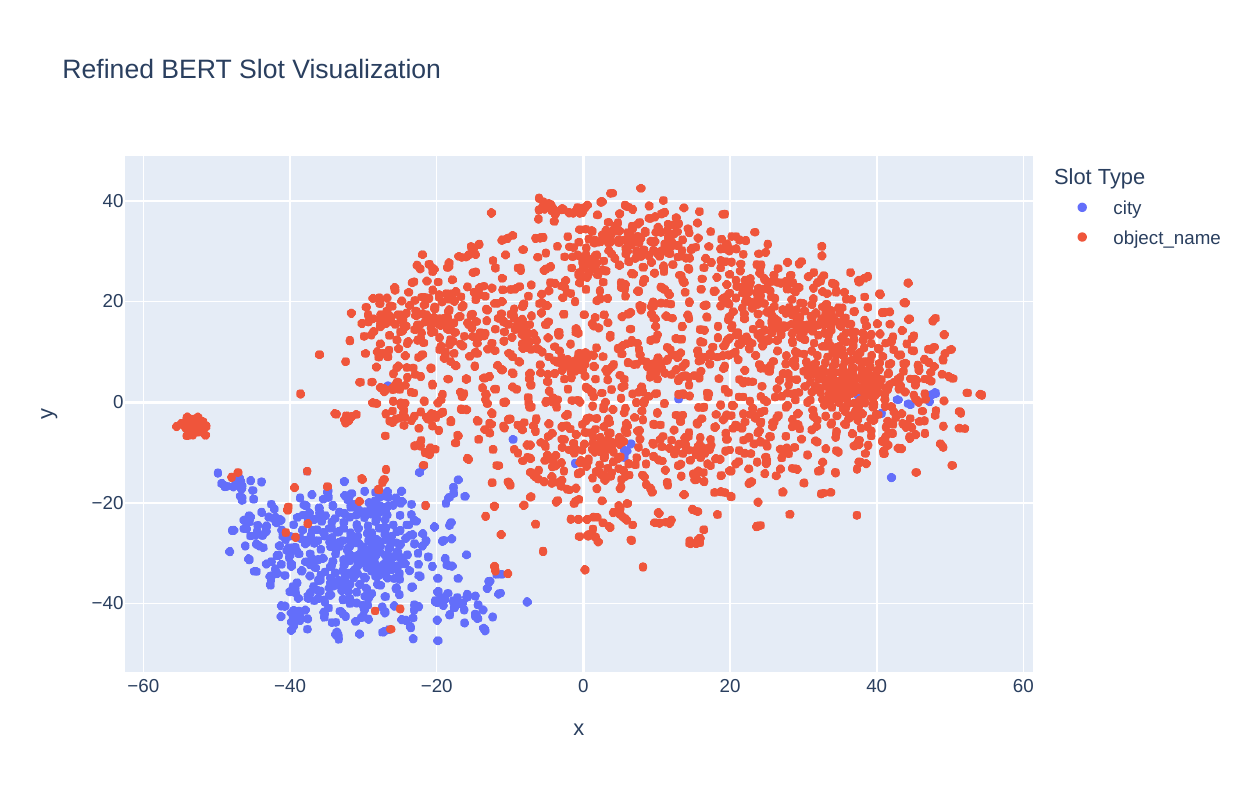} }
      \label{subfig:refined_test}
    }%
    \vspace*{-0.2cm}
    \caption{Slot Value Representation Visualization of the raw  original pre-trained BERT and raw Refined BERT via SI on sample slot types from training set SNIPS\_P1 (\protect\subref{subfig:pre_train}, \protect\subref{subfig:refined_train}) and testing set SNIPS\_P2 (\protect\subref{subfig:pre_test}, \protect\subref{subfig:refined_test}). \textcolor{blue}{Blue} and \textcolor{red}{Red} denotes slot values from randomly sampled ground truth slot types.}
    \label{fig:vis_emerge}%
    \vspace*{-0.60cm}
\end{figure*}
\vspace*{-0.2cm}
\paragraph{Sentence-level Supervision (SentCL)} On the sentence level, besides the commonly aligned phrases (i.e. \textit{add tune to} vs \textit{add rupee to}), the model recognizes corresponding playlists in anchor and positive samples (i.e. \textit{black metal playlist} vs \textit{ultra metal playlist}) and assign competitive similarity score between them. On the other hand, potential relevant noun phrases (i.e. ultra metal playlist (\textit{playlist}) and andy silvester sound track (\textit{sound track})) between anchor and negative samples are assigned low similarity score. This showcases the model's capability in (1) correctly recognizing and bringing the important slot phrases in positive-anchor pair closer together, (2) reducing the importance of potential relevant slot phrases across samples with different intents. The Similarity Matrix presented in Figure \ref{subfig:dist_pos_anchor} also indicates the strong segment alignment between positive and anchor samples as the diagonal cells receive higher similarity score than most of the other cells within the same column or row. 
\paragraph{Qualitative Case Study}
Additional Case Studies presented in Figure \ref{fig:case} demonstrate the effectiveness of our proposed framework in capturing slot phrases. Despite the imperfect segmentations, \textit{Ours} captures phrases closer to the ground truth slot labels than other \textit{Comparable} baselines. In fact, our identified phrases ``spirit touches ground'' and ``leche con chocolate list'' are exact matches for the golden slot labels. Our proposed multi-level CL refining mechanism is also shown to correct mistakes of the original model. (from ``by phil'' in~\textit{Ours (w/o CL)} to ``phil och'' in~\textit{Ours (with CL}).  

\begin{table}[bt]
\centering
\caption{Evaluation of SF task over 3 runs on Emerging Intents in SNIPS\_P2 and ATIS\_P2 datasets.}
\vspace*{-0.3cm}
\resizebox{\columnwidth}{!}{%

\begin{tabular}{|c||c|c|c||}
\hline 
  & \multicolumn{3}{c||}{\textbf{SNIPS\_P2}} 
  \\
 \hline
 & S-P & S-R & S-F1 \\
 \hline
Original BERT & 14.11 $\pm$ 0.47 & 17.78 $\pm$ 0.82 & 15.73 $\pm$ 0.62 
\\
\hline
Refined BERT & \textbf{15.08 $\pm$ 0.48} & \textbf{19.61 $\pm$ 0.23} & \textbf{17.05 $\pm$ 0.38} 
\\
\hline
\end{tabular}%
}
\\

\resizebox{\columnwidth}{!}{%
\begin{tabular}{|c||c|c|c||}
\hline 
  & \multicolumn{3}{c||}{\textbf{ATIS\_P2}} 
  \\
 \hline
Original BERT 
& 66.67 $\pm$ 0.82 & 63.35 $\pm$ 1.35 & 64.96 $\pm$ 0.74 
\\
\hline
Refined BERT 
& \textbf{70.12 $\pm$ 0.85} & \textbf{63.64 $\pm$ 0.48} & \textbf{66.72 $\pm$ 0.66} 
\\
\hline
\end{tabular}%
}
\label{tab:downstream}
\vspace*{-0.6cm}
\end{table}

\vspace*{-0.2cm}
\subsection{Generalization towards Emerging Intents}
 \vspace*{-0.2cm}
\paragraph{Visual Representation}
We first visualize the representations of two randomly sampled slot types produced by the raw original BERT and our Refined BERT (via SI objectives). As observed in Figure \ref{fig:vis_emerge}, our Refined BERT clusters the representations of samples with the same slot types for both training and testing sets more effectively than the original BERT in the embedding space, leading to far clearer separation boundaries between the sampled slot types. For Train Slots, embeddings of slot values from each slot type are nearly disentangled, implying our Refined BERT is capable of recognizing slot types without explicit slot training objectives and token-level label access. In addition, when applied to new intents and slots in P2 dataset, our SI framework produces refined BERT with better semantic representations for tokens from the same slot types as observed in Figure \ref{subfig:pre_test},\ref{subfig:refined_test}.


\paragraph{Quantitative Evaluation} 
As observed in Table \ref{tab:downstream}, when generalized to emerging intents and slots, our Refined BERT outperforms the traditional BERT while fine-tuning on both datasets in all slot evaluation metrics. This showcases the generalization capability of our model across different sentence-level intent labels. In addition, the consistent improvement in SF evaluation implies that SI training objectives via UPL and CL refinement provide more guidance to the PLM for the downstream token-level task without explicit training objectives and label requirements.  

\vspace*{-0.2cm}
\section{Conclusion}
\vspace*{-0.2cm}
In our work, we propose the study of token-level Slot Induction (SI) via an Unsupervised Pre-trained Language Modeling (PLM) Probing in conjunction with Contrastive Learning (CL) objectives. By leveraging both unsupervised signals from PLM and sentence-level signals from intent labels via CL objectives, our proposed framework not only achieves competitive performance in comparison with other unsupervised phrasal segmentation baselines but also bridges the gap in performance with \textit{Upper Bound} methods that require additional token-level labels on two NLU benchmark datasets. We also demonstrate that our proposed SI training is capable of refining the original PLM, resulting in more effective slot representations and benefiting downstream SF tasks when generalized towards emerging intents. Further studies of better exploitation of full-depth segmentation trees, enhanced segment augmentation mechanisms and better semantic alignment extraction between slots and intents are promising directions for our future work. We also seek to extend the current SI studies beyond English and towards multilingual NLU systems. \cite{nguyen2019cross, qin-etal-2022-gl, nguyen-etal-2023-enhancing}

\label{sec:bibtex}

\section*{Limitations}
Our proposed framework assumes a fixed hyperparameter depth $d$ for UPL segmentation tree. In other words, only segments extracted at the depth $d$ are considered for CL objectives. $d$ is tuned with each dataset's validation set. However, as our main objective is to investigate the effects of UPL and CL objectives, we leave the full tree exploitation as future extensions for our work.

Secondly, the goal of our SI is to identify the slot phrase boundaries. The label type predictions for recognized slot phrases are beyond the scope of our investigation. Therefore, direct end-to-end evaluation of SI in mitigating slot label scarcity issues cannot be directly evaluated. Our rationale for dividing the task into 2 separate steps (i.e. slot boundary induction and slot label prediction) is as follows: As the complete SI is a complex task, breaking it down not only allows for direct and focused evaluation of the proposed framework's contribution at individual steps but also minimizes error propagation from intermediate steps to a single end-task metric. This rationale is further supported by our empirical study in Section \ref{sec:result}. The proposed \textit{USSI} whose objective unifies both aforementioned steps underperforms \textit{Ours(w/o CL)} and \textit{Ours(full)} when evaluated at the slot boundary induction step.

\section*{Acknowledgement}
This work is supported in part by NSF under grants III-1763325, III-1909323,  III-2106758, and SaTC-1930941.

We would like to acknowledge the use of the facilities of the High Performance Computing Division and High Performance Research and Development Group at the National Center for Atmospheric Research and the use of computational resources
(doi:10.5065/D6RX99HX) at the NCAR-Wyoming Supercomputing Center provided by the National Science Foundation and the State of Wyoming, and supported by NCAR’s Computational and Information Systems Laboratory.

\bibliography{anthology,custom}
\bibliographystyle{acl_natbib}
\newpage 
\appendix
\section{Slot Induction Baselines}
For fair comparisons across all baselines, we leverage BERT \cite{devlin2019bert} as the backbone PLM architecture (if applicable). 
\label{sec:ap_base}
\subsection{Upper Bound Baselines}
\label{sec:ap_base_up}
\noindent$\bullet$ \textbf{Joint BERT FT}: Fully Supervised   Joint Sequence Labeling and Sentence Classification model is trained on top of fine-tuning BERT embeddings with available golden training slot and intent labels.\\
\noindent$\bullet$ \textbf{SpaCy}~\cite{Honnibal_spaCy_Industrial-strength_Natural_2020}: Industrial-strength NLP tagging methodology that leverages pre-trained NP chunking model. \\
\noindent$\bullet$ \textbf{FlairNLP}~\cite{akbik2018coling}: Neural Language Modeling in junction with pre-trained Sequential Labeling (NER and POS).   
    
\subsection{Comparable Baselines}
\label{sec:ap_base_comp}
\noindent$\bullet$ \textbf{Dependency Parsing} (Right/Left-branching (RB/LB): Parameter-free methods for sentence segmentation. Result from the best depth is reported.\\
\noindent$\bullet$ \textbf{AutoPhrase}~\cite{shang2018automated}: Statistical phrase tagging method utilizing high quality massive corpus as additional Knowledge Base (KB). \\
\noindent$\bullet$ \textbf{UCPhrase}~\cite{10.1145/3447548.3467397}: Phrase tagging method leveraging co-occurrence word frequency and PLM attention maps.  \\
\noindent$\bullet$ \textbf{USSI}~\cite{yu2022unsupervised}: Unsupervised Slot Schema Induction method leveraging attention distribution of PLM and additional constraints from Probabilistic Context-free Grammar (PCFG) \cite{Kim2020Are}. For completeness, additional experiments in leveraging the proposed in-domain training objectives with SpanBERT PLM \cite{joshi2020spanbert} are provided in Appendix \ref{sec:spanbert}.  \\
\noindent$\bullet$ \textbf{Ours (w/o CL)}: 
Fixed UPL is directly used for inference without additional CL refinement. Same depth $d$ is used as our proposed model \textbf{Ours (full)} and its variant \textbf{Ours (w/o SentCL)}. \\
\noindent$\bullet$
\textbf{Ours (w/o SentCL)}: Our model variant that is trained only with SegCL objectives ($\mathcal{L}_{s}$). The model does not leverage sentence-level intent label information (SentCL) during training. 
\begin{table}[tb]
\centering
\caption{Hyperparameters for SNIPS and ATIS datasets (SI task)}
\vspace*{-0.2cm}
\resizebox{0.8\columnwidth}{!}{%
\begin{tabular}{|c||c|c|c|c|c|c||}
\hline 
 & d & $\beta$ & $\tau_{s}$ & $\tau_{d}$ & $\delta$ & $\gamma$\\
\hline
SNIPS & 3 & 0.2 & 0.1 & 0.05 & 0.3 & 0.7   \\
\hline 
ATIS & 4 & 0.2 & 0.05 & 0.1 & 1.0 & 0.2 \\
\hline
\end{tabular}%
}
\vspace*{-0.5cm}
\label{hyperparameter}
\end{table}
\section{Slot Induction Implementation (P1)}
\label{sec:ap_hyper}
We train our proposed SI model with batch size of 16, learning rate 1e-5 for 10 epochs. The remaining hyperparameters for individual datasets are reported in Table \ref{hyperparameter} respectively for SI task. We tune our hyperparameters based on each dataset's P1 validation set via grid search for $\beta,\tau_{s}, \tau_{d}, \delta, \gamma$, except for $d$. For depth $d$, we conduct inference of PLM probing (i.e. Ours (w/o CL)) on P1 validation sets and select $d$ with the highest H-Mean performance. The same depth $d$ is used consistently across different variants of our proposed framework in the empirical study. Our reported results are reported based on 3 runs with different seeds.      

\section{Slot Filling Implementation (P2)}
\label{sec:ap_sf}
As the objective of SF is to compare different BERT models (i.e. Original BERT vs Refined BERT via SI objectives), we keep the Sequence Labelling architecture simple and similar between the two models. Specifically, we stack the traditional CRF layer \cite{lafferty2001conditional} on top of the corresponding BERT models. The overall model is fine-tuned on SF task with available training slot labels in P1\ training data. The model is fine-tuned with batch size of 16, learning rate of 0.01 for CRF and Linear layer, BERT learning rate of 1e-5 for 10 epochs. The testing results (Table \ref{tab:downstream}) are reported on P2 of each dataset as an average over 3 runs. Both training and inference for Appendix \ref{sec:ap_hyper} and \ref{sec:ap_sf} are conducted on NVIDIA Titan RTX GPU.

\begin{table}[tb]
\centering
\caption{Ablation study of SpanBERT PLMs with in-domain training objectives on SNIPS and ATIS datasets in terms of H-Mean over 3 runs. $^{\P}$ denotes models that do not require random initializations.}
\vspace*{-0.3cm}
\resizebox{0.9\columnwidth}{!}{%
\begin{tabular}{|c||c|c|c|c||}
\hline 
  & \multicolumn{1}{c|}{\textbf{SNIPS}} & \multicolumn{1}{c|}{\textbf{ATIS}} \\
 \hline
SpanBERT $^{\P}$ & 43.15 & 35.05 \\
\hline
USSI \cite{yu2022unsupervised}
& 48.61 $\pm$ 0.69 & 36.63 $\pm$ 1.93 \\
\textbf{Ours (SpanBERT w CL)}
& \textbf{53.25 $\pm$ 0.29} & \textbf{40.07 $\pm$ 2.34} \\
\hline
\end{tabular}%
}
 \vspace*{-0.5cm}
\label{tab:spanbert}
\end{table}

\section{SpanBERT-based Model}
\label{sec:spanbert}
\citet{yu2022unsupervised} proposed additional self-supervised in-domain training on Task-oriented Dialogue datasets. For fair comparisons with \cite{yu2022unsupervised}, we conduct additional studies training the same backbone SpanBERT PLM architecture \cite{joshi2020spanbert} with their proposed self-supervised in-domain training objectives on our training SNIPS\_P1 and ATIS\_P1 datasets and report test results in Table \ref{tab:spanbert}. To evaluate the effectiveness of our multi-level CL objectives, in Table \ref{tab:spanbert}, \textbf{Ours (SpanBERT w CL)} follows the induction mechanisms proposed by \citet{yu2022unsupervised} instead of UPL mentioned in Section \ref{subsec:upl}. The only difference between \textbf{Ours (SpanBERT w CL)} and USSI is our proposed multi-level CL objectives    

As demonstrated in Table \ref{tab:spanbert}, \textbf{Ours (SpanBERT w CL)} achieves consistent improvements over USSI on both SNIPS and ATIS datasets (4.64\% and 3.44\% respectively) under the same training architecture and in-domain training objectives. This observation implies the effectiveness of our multi-level CL objectives (SegCL and SentCL). 



\end{document}